\documentclass[10pt,twocolumn,letterpaper]{article}

\usepackage{cvpr}
\usepackage{multicol}
\usepackage{times}
\usepackage{epsfig}
\usepackage{graphicx}
\usepackage{amsmath}
\usepackage{amssymb}
\usepackage[utf8]{inputenc} % allow utf-8 input
\usepackage[T1]{fontenc}    % use 8-bit T1 fonts
\usepackage{url}            % simple URL typesetting
\usepackage{booktabs}       % professional-quality tables
\usepackage{amsfonts}       % blackboard math symbols
\usepackage{nicefrac}       % compact symbols for 1/2, etc.
\usepackage{microtype}      % microtypography
\usepackage{algorithm}
\usepackage{graphicx}
\usepackage{wrapfig}
\usepackage{subcaption}
\usepackage[noend]{algpseudocode}
\usepackage{caption}
\usepackage{placeins}
\usepackage{booktabs}
\usepackage{comment}    % temporary

\usepackage[pagebackref=true,breaklinks=true,letterpaper=true,colorlinks,bookmarks=false]{hyperref}

\cvprfinalcopy % *** Uncomment this line for the final submission

\def\cvprPaperID{2712} % *** Enter the CVPR Paper ID here

\ifcvprfinal\fi
\begin{document}

% Draft title
%\title{Point R-CNN?: Point Cloud Based Multi-Person 3D Pose Estimation using Multi-Cameras ???  View Invariant??}
 \title{View Invariant Human Body Detection and Pose Estimation from Multiple Depth Sensors}

% The \author macro works with any number of authors. There are two
% commands used to separate the names and addresses of multiple
% authors: \And and \AND.
%
% Using \And between authors leaves it to LaTeX to determine where to
% break the lines. Using \AND forces a line break at that point. So,
% if LaTeX puts 3 of 4 authors names on the first line, and the last
% on the second line, try using \AND instead of \And before the third
% author name.

\author{ Walid Bekhtaoui, Ruhan Sa, Brian Teixeira,  Vivek Singh, Klaus Kirchberg, \and Yao-jen Chang Ankur Kapoor\\Siemens Healthineers, Digital Services, Digital Technology \& Innovation,
Princeton, NJ, USA}

% vectors in bold
\renewcommand{\vec}[1]{\mathbf{#1}}
\date{}

\twocolumn[{
\renewcommand\twocolumn[1][]{#1}%
\maketitle
\vspace{-3em}
%\begin{figure*}
    \begin{center}
    \includegraphics[width=0.9\linewidth]{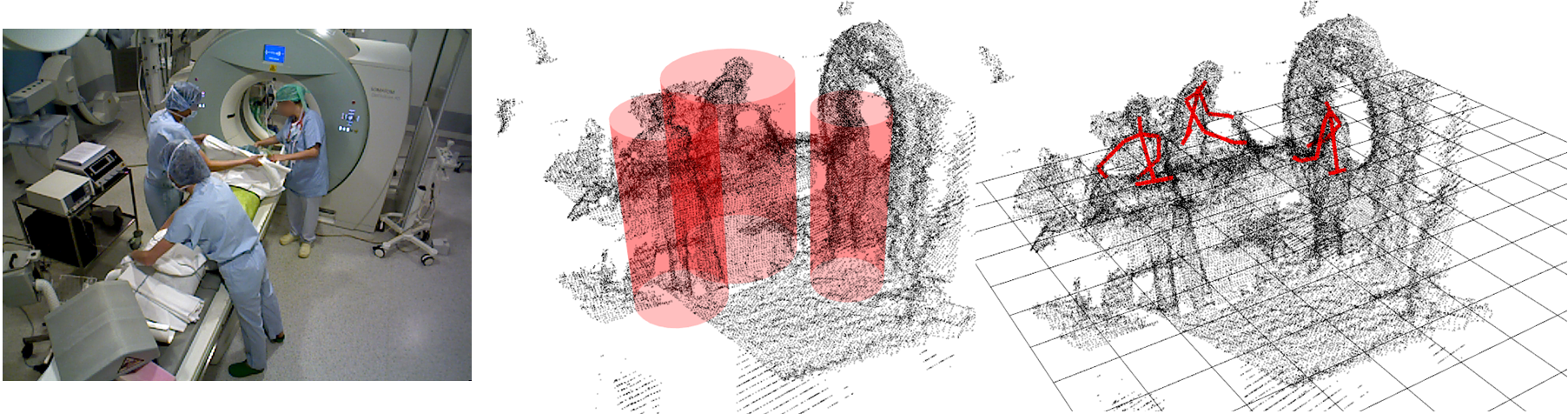}
     \captionof{figure}{\textbf{Point R-CNN.} Scene from the Multi View Operation Room (MVOR) dataset. (a) shows a view from RGB camera (used only for visualization), (b) and (c) show the example of person detection and pose detection on the merged point cloud from 3 depth cameras.}
    \label{fig:teaser}
    \end{center}
%\end{figure*}
}]

\begin{abstract}
Point cloud based methods have produced promising results in areas such as 3D object detection in autonomous driving. However, most of the recent point cloud work focuses on single depth sensor data, whereas less work has been done on indoor monitoring applications, such as operation room monitoring in hospitals or indoor surveillance. In these scenarios multiple cameras are often used to tackle occlusion problems.
%One of the challenges in multi-camera scenarios is  bring other challenges, such as fusion of data from multiple data sources and individual camera failures. To this end, 
We propose an end-to-end multi-person 3D pose estimation network, \emph{Point R-CNN}, using multiple point cloud sources. We conduct extensive experiments to simulate challenging real world cases, such as individual camera failures, various target appearances, and complex cluttered scenes with the CMU panoptic dataset and the MVOR operation room dataset.
Unlike most of the previous methods that attempt to use multiple sensor information by building complex fusion models, which often lead to poor generalization, we take advantage of the efficiency of concatenating point clouds to fuse the information at the input level. In the meantime, we show our end-to-end network greatly outperforms cascaded state-of-the-art models. \footnote{This feature is based on research, and is not commercially available. Due to regulatory reasons its future availability cannot be guaranteed.}
\end{abstract}

%\begin{abstract}
%Detecting 3D objects in real world conditions is a challenging problem in Computer Vision. 
%Recent works explored this topic in the context of autonomous driving, which
%often relies on a single viewpoint. However, most indoor monitor applications such as surveillance
%or operation room monitoring in hospitals use several viewpoints for representing the scene.
%While this configuration surely helps address object occlusions, it introduces other challenges
%such as camera registration and synchronization. In this paper, we introduce ``Point R-CNN'', an end
%to end multi-view and multi-person human segmentation and 3D pose estimation. Unlike previous works,
%our approach trains end-to-end and does not use complex fusing models which often suffers from poor generalization.
%We demonstrate the effectiveness of our solution on two challenging datasets: CMU panoptic \cite{joo2018total} 
%a controlled dataset of human pose from 10 RGB-D sensors and MVOR \cite{mvor} a medical dataset of 
%complex cluttered scenes, where our end-to-end method outperforms previous cascaded models.

% \footnote{This feature is based on research, and is not
%  commercially available. Due to regulatory reasons its future
%  availability cannot be guaranteed.}

%\end{abstract}

\section{Introduction}
Point cloud analysis has been studied widely due to its important applications in autonomous driving\cite{zhou2017voxelnet,qi2017frustum,ku2017joint}, augmented reality\cite{park2008multiple}, and medical applications\cite{singh2017darwin}. However, most of the point cloud based work has been focusing on applications with only one depth sensor, or multiple sensors facing outwards, covering non-overlapping regions. Less work has been reported that tackles the 3D object detection problem in in-door multi-camera settings, which is a very common and important scenario in applications such as operating room monitoring\cite{kadkhodamohammadi2017multi}, indoor social interaction studies\cite{joo2018total}, indoor surveillance \etc.

Using multiple cameras reduces the amount of occlusions, but fusing information from multiple sensors is very challenging and still an open problem in the community. Traditionally fusing 2D and or 3D images coming from multiple sensors is handled by complex models that either process each sensor separately and fuse the decision at a later stage \cite{cho2014multi, gupta2015aligning,tulsiani2015viewpoints}, or fuse the information earlier on feature level\cite{xu2017pointfusion}. Such algorithms tend to suffer from poor generalization due to the complexity of the model and heavy assumptions. In comparison, we argue that using multiple sourced point clouds is a more straightforward and natural alternative, which provides better generalization under various challenging real world scenarios.

%\begin{figure}[htb]
%\includegraphics[width=\linewidth]{imgs/framework.png}
%\caption{Overview of the Point R-CNN framework.}
%\label{img_framework}
%\end{figure}

In this work we study the multi-person 3D pose estimation problem for indoor multi-camera settings using point cloud and propose an end-to-end multi-person 3D pose estimation network, ``Point R-CNN''. %, as shown in \autoref{img_framework}%.
We tested our method by simulating scenarios such as camera failures and camera view changes on the CMU panoptic dataset\cite{joo2018total}, which was collected in the CMU Panoptic studio to capture various social interaction events with 10 depth cameras. We further test our method on the challenging real world dataset MVOR\cite{mvor}, which was captured in hospital operation rooms with 3 depth cameras. The experiment demonstrates the robustness of the algorithm for challenging scenes and shows good generalization ability on multi-sensor point clouds. Furthermore, we show that the proposed end-to-end network outperforms the baseline cascaded model by a large margin.

The contributions of our work are as follows:

\begin{enumerate}
\item We propose to use point cloud as the only source for fusing data from multiple cameras and show that our proposed method is efficient and generalizes well to various challenging scenarios.
\item We propose an end-to-end multi-person 3D pose estimation network, Point R-CNN, based solely on point clouds. Through extended experiments, we show the proposed network outperforms the cascaded state-of-the-art models.
\item We present extensive experimental results simulating challenging in-door multi-camera application problems, such as repeated camera failures and view changes. 
\end{enumerate}

\section{Related work}

\subsection{Point cloud based approaches}
Processing point cloud data is challenging due to its unstructured nature. Typically, in order to use Convolutional Neural Network (CNN) like methods, point clouds are usually pre-processed and projected to some ordered space, such as in \cite{wu20153d,riegler2017octnet,jampani2016learning}.
Projecting point clouds is also helpful for more complex localization tasks such as 3D object detection. For example, Zhou \etal\cite{zhou2017voxelnet} proposed to use voxelized point clouds to detect 3D objects using a Region Proposal Network, which simultaneously produces voxel class labels and regression results for bounding box ``anchors''.

Alternatively, more efforts have been put into directly using point clouds. For example, Qi \etal proposed PointNet\cite{qi2017pointnet} and PointNet++\cite{qi2017pointnet++} to classify and segment point clouds in their native space. These networks are the building blocks for many later point cloud based algorithms. Recently, Ge \etal\cite{ge2018hand} proposed to directly use point clouds for hand pose estimation. We further discuss this approach in section \ref{sec:pose}.

Besides the above mentioned trends, on-the-fly point cloud transformations have also been explored. Su \etal\cite{su2018splatnet} proposed using Bilateral Convolutional Layers to filter the point cloud and further process the data without explicitly pre-processing the input point cloud. Li \etal\cite{li2018pointcnn} proposed to use $\chi$-transform to gradually transform the point clouds into a higher order representation without pre-processing steps. 

Our method is inspired from the first two approaches to detect people in ordered 3D space
using voxelized input, as well as detecting human body joints on the segmented point cloud.
%In this paper, we bring in the best of first two approaches to locate persons in 3D ordered space using voxelized input and further regress the joints directly on the detected person point cloud. 

\subsection{Multi-sensor applications}
As discussed earlier, combining the information of multiple sensors is challenging. Conventionally the information is fused at a later stage, either by fusing the decision or fusing the feature space. Hedge \etal\cite{hegde2016fusionnet} proposed to use one network per sensor and fuse the result at a later stage to perform object recognition. Xu \etal proposed PointFusion\cite{xu2017pointfusion} to fuse point cloud features and 2D images to detect 3D bounding boxes.

However, complex fusion models present weaknesses in terms of generalization. For example, in indoor surveillance systems, the number of cameras and their position vary widely making it challenging to generalize. Hence we argue that combining multiple point cloud sources is a more natural and effective alternative where the effort of fusing information is very low in comparison. Furthermore, in case of camera failures, the structure of the input does not change, only the density, \ie number of points does. This can easily be accounted for by training the point cloud network on input cloud with variable density.

\subsection{3D human pose estimation}
\label{sec:pose}
3D human pose estimation is a challenging problem. Most recent works focused on using depth images or combining 2D and 3D information to detect landmarks for a single or for multiple persons\cite{sarafianos20163d,yang20183d,rhodin2018learning,kadkhodamohammadi2018generalizable}.

Rhodin \etal\cite{rhodin2018learning} proposed to use weakly-supervised training methods to circumvent the annotation problem. This is achieved by assuming consistency in the pose across different views. During testing the pose is estimated based on a single camera input. More recently, Moon \etal\cite{moon2018v2v} proposed to use voxelized depth images to detect 3D hands and estimate single human pose from depth images. Haque \etal \cite{haque2016pose3d} describe a view point invariant 3D human pose estimation network, where the input depth image is either a top view or a front view. They refine the landmark position using a Recurrent Neural Network and tested view transfer by training on the front view and testing in on the side view. While this relates to our problem and presents encouraging results, the assumption of having fixed viewpoints does not apply for our application.

\section{Approach}
The work we are presenting is built upon the VoxelNet paradigm\cite{zhou2017voxelnet}.
Our end to end framework can be split into two parts, (1) instance detection and (2) instance processing.

\begin{figure*}[htb]
\centering
\includegraphics[width=\linewidth]{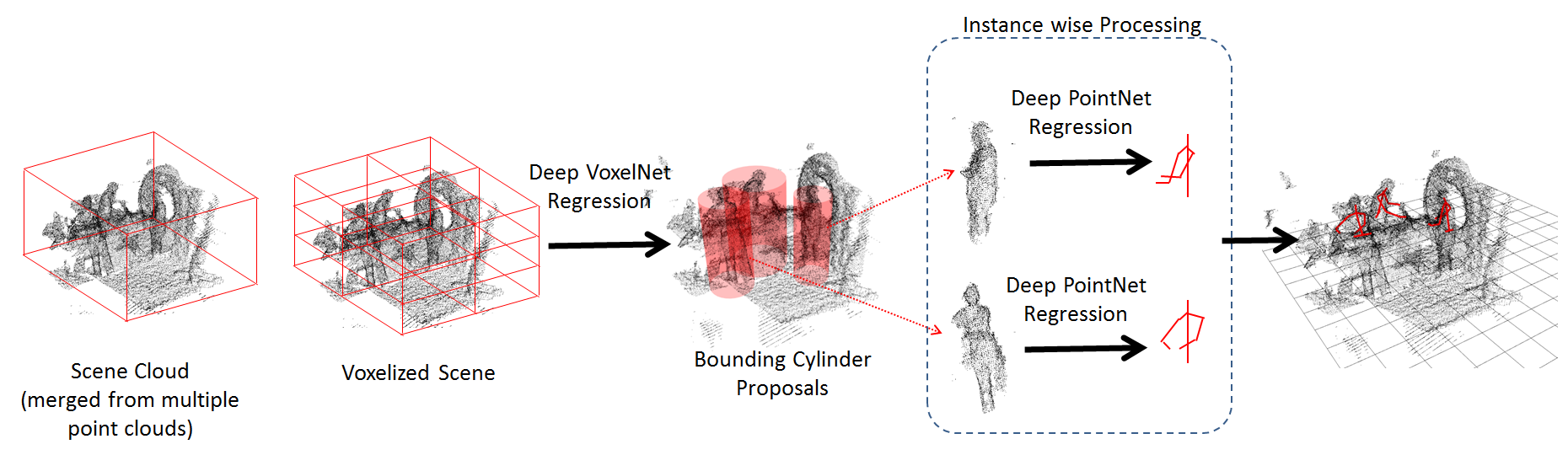}
\caption{Overview of the Point R-CNN framework.}
\label{fig:img_framework}
\end{figure*}

\subsection{Architecture} 
Our framework is outlined in \autoref{fig:img_framework}.
The architecture can be split into several modules which are (1) per-voxel feature extraction, (2) voxel features aggregation, (3) instance detection, and finally (4) instance-wise processing, \ie in our case point to point regression. In the following sections we describe these modules in detail.

%\begin{figure*}[htb]
%    \includegraphics[width=0.9\linewidth]{framework22.pdf}
%    \caption{
%    The framework described here take as input a list of points which correspond to the points of our 3D scene. Step (1) shows how we efficiently assign each point to a voxel. In this example, we consider a scene with 4 voxels where each can only have 3 points in them.  The (2) let us go from a set of points to a voxel grid.
%  (3) use the bounding cylinder regressed to extract the points and create a mini batch to feed to PointNet\cite{qi2017pointnet}. Here we used a $L'$ but we could use $L$ as well.
%  }
%  \label{fig:fw}
%\end{figure*}

\subsection{Input preprocessing}

%\paragraph{Voxel Partition and grouping.}
\begin{figure}[htb]
    \centering
    \includegraphics[width=\linewidth]{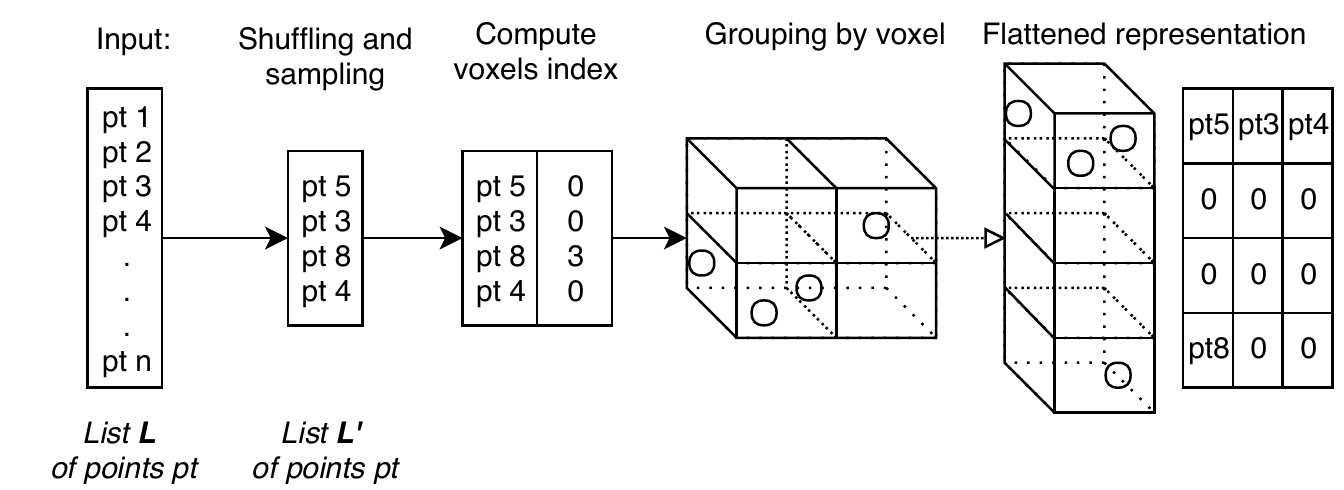}
    \caption{The input of our architecture is the concatenation of the point clouds from several sensors. During this step we associate each point of our scene to a voxel which allows us to easily access a point and his neighborhood}
  \label{fig:input_proc}
\end{figure}
The input to our algorithm is the unstructured point cloud $\vec{P}=\bigcup \vec{P}_i$, where $\vec{P}_i$ denotes the point cloud acquired from sensor $i$. The point clouds acquired from all the sensors are assumed to be time-synchronized and registered within the same world coordinate system. We further assume our world coordinate system to be axis aligned with the ground plane with the $y$-axis being the ground normal.

As a first step we define an axis-aligned cuboid working space resting on top of the ground plane. All points outside this volume are discarded at this time.

In order to reduce the impact of variable point cloud densities across the scene and to speed up processing we down-sample $\vec{p}$ using a voxel grid filter to our working space. This filter merges all points that fall within one voxel into their centroid such that after filtering each voxel contains at most one point.

We then subdivide the working space into a different regular axis-aligned grid of (larger) voxels as follows. The origin of our working space is denoted by $\vec{o}$, the dimensions of each voxel $v_x$, $v_y$ and $v_z$, and the number of voxels along each axis $N_x$, $N_y$ and $N_z$. We also choose the number $T$ of points to be considered per voxel, in our case we chose $T=64$.

Each point $\vec{p}$ of the point cloud is now assigned to the voxel it falls in, denoted by the directional voxel indices
$id_x=\lfloor{(p_x-o_x)/v_x}\rfloor$,
$id_y=\lfloor{(p_y-o_y)/v_y}\rfloor$ and
$id_z=\lfloor{(p_z-o_z)/v_z}\rfloor$, where $0 \leq id_x < N_x$ \etc .

% In the next step we subsample the point cloud by randomly picking $min(N_v * T, N_{pts})$ points, where $N_v=N_x*N_y*N_z$ is the total number of voxels, $T$ denotes the fixed target number of points for each voxel, and $N_{pts}$ the total number of remaining points in the point cloud.

% The reason for this subsampling is twofold. Reducing the number of points decreases the required memory and computation time. In addition it reduces the impact of highly variable point densities in different regions. 

% In order to do the grouping in an efficient manner, we use the fact that we know how many voxel we have in our scene and that our grid is aligned with $x$, $y$ and $z$ axis. With this information, we can easily compute the index of the voxel in which a point belongs to. Let consider the point $pt$ in our scene, then to compute the voxel index we have:

% \begin{align}
% id_x = \frac{- ((pt_x - p_{0x}) \bmod v_x) + (pt_x - p_{0x})}{v_x} \nonumber \\
% id_y = \frac{- ((pt_y - p_{0y}) \bmod v_y) + (pt_y - p_{0y})}{v_y} \nonumber \\
% id_z = \frac{- ((pt_z - p_{0z}) \bmod v_z) + (pt_z - p_{0z})}{v_z} \nonumber
% \end{align}

%Then to go from those 3 indexes to a unique one we apply:

The voxel grid is then flattened by assigning a linear index to each voxel via 
$$
id = id_z + id_y * N_z + id_x * N_z * N_y .
$$

%With this $id$ we can then easily discard the point outside of our scene, the points which are not in the area we are interested in. This can be done by checking if the $id$ is not between $0$ and our total number of voxel $N_v$ which is $n_x * n_y * n_z$ .
After this grouping, each point is assigned the $id$ of the corresponding voxel.

Using the $id$ we previously computed for each point, we can find the list of unique $id$s of every voxel containing at least one point. 
Since we already sampled and shuffled the whole point cloud, we just have to take for each voxel the $T$ first points with this $id$ and put those $T$ points in a tensor of size $(3, T, N_v)$, 3 being our input dimension and $N_v$ being the total number of voxel in the scene $N_x*N_y*N_z$.

Instead of using the world coordinate $\vec{x}$ of each points we use the relative position $\vec{x}'$ within its corresponding voxel, \ie
$$
\vec{x}'=\vec{x}-\left(\vec{o}+\begin{pmatrix}N_xv_x\\N_yv_y\\N_zv_z\end{pmatrix}\right) .
$$
This prevents the network from learning global aspects of the scene as opposed to the desired local structures,

%Speaking about the input dimension, we could have used the coordinates of the points to train our network, but in order to prevent our network from learning the scene disposition, we instead replace each point coordinate by its difference to the voxel it belongs to.
We pad with zeros the voxel which do not have enough points.

\subsection{Instance detection} 
Now that our scene is defined, we can regress the bounding cylinder for each instance (\ie person) in the scene. Our approach to do this is inspired by previous work on 2D instance detection and segmentation \cite{girshick2015fastrcnn, he2017maskrcnn}.

\paragraph{Per-voxel feature extraction.}
\begin{figure}[htb]
    \centering
    \includegraphics[width=\linewidth]{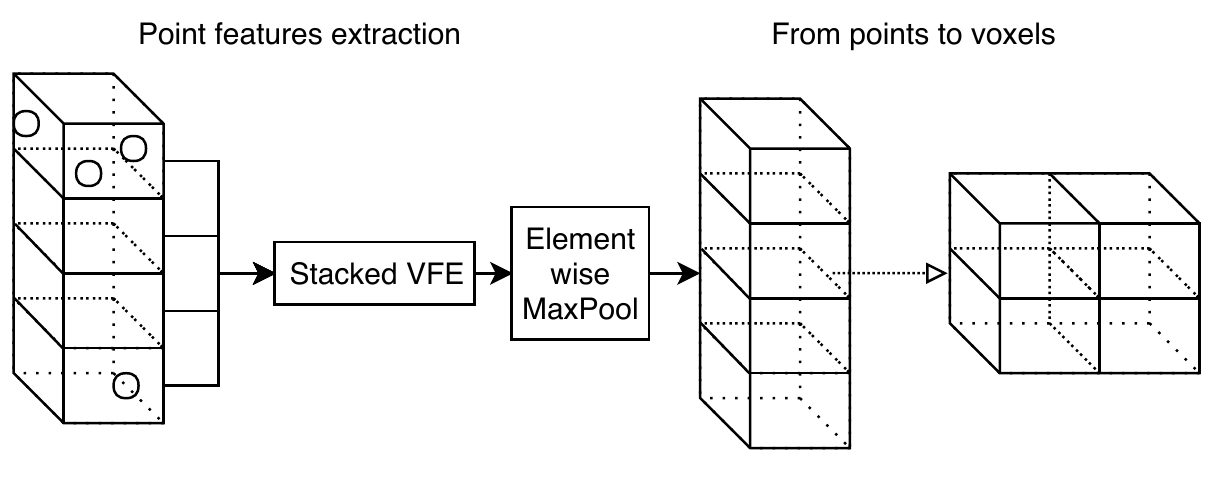}
    \caption{Per-voxel feature extraction pipeline.
    Each point is processed using VFE\cite{zhou2017voxelnet} and get some information from the point in the same voxel. Finally, element-wise max pooling let us retrieve a feature vector per voxel.}
  \label{fig:point_feat}
\end{figure}

The first part of our architecture is using 2 stacked \emph{voxel feature encoding} (VFE) layers \cite{zhou2017voxelnet} in order to learn a voxel-wise set of features which is invariant to point permutation.
Those VFE layers are efficiently implemented as 2D convolutions with kernels of size 1.
We use a similar notation as in \cite{zhou2017voxelnet} for the VFE layers.
To represent the i-th VFE layer, we use VFE-i$(c_{in},c_{out})$ with $c_{in}$ and $c_{out}$ being respectively the dimension of the input features and the dimension of the output features of the layer. As in \cite{zhou2017voxelnet},
the VFE layer will transform its input into a feature vector of dimension $c_{in}\times(c_{out}/2)$ before doing the point-wise concatenation, which then yields the output of dimension $c_{out}$.

The VFE are VFE-1(3, 32) and VFE-2(32, 64) followed by a fully connected layer of input and output 64 right before the element-wise max pooling. 
Having the VFE instead of solely using PointNet \cite{qi2017pointnet} helps add  neighborhood information to the points (defined by being every point in this voxel).
After the max pooling of the last layer, the output size is $(64, 1, N_v)$. 
We can thus reshape our output in order to retrieve a 3D image of size $(64, N_x, N_y, N_z)$.

In this work we process every voxel. However, this could be sped up as shown in \cite{zhou2017voxelnet} by only processing non-empty voxels.

% \paragraph{Voxel feature aggregation.}
% \begin{figure}[htb]
%     \includegraphics[width=\linewidth]{voxels_feat.pdf}
%     \caption{Using the extracted feature from the previous step, we use a 3D DenseUNet \cite{denseunet} to aggregate features from the scene at various resolutions. The output of the DenseUNet is then fed into two parallel branches to do the detection and regression at the voxel level.}
%   \label{fig:voxel_feat}
% \end{figure}

In order to aggregate information from the scene and learn multi-scale features from our scene, we use a DenseUNet \cite{denseunet}.
The first step makes our network invariant to point permutation. Then, when working on voxels instead of point clouds, we go from an unstructured representation to a structured one, which is easier to apply classic deep learning methods on. By doing those two steps we provide additional neighborhood information to each point at voxel and scene level. This is more efficient than processing the whole point cloud and looking for k-closest neighbor, and it gives better results than just processing the whole point cloud as a set of patches.

\paragraph{Instance detection.}
After the feature aggregation part we take the output and feed it to two parallel heads: one for the per-voxel classification (doing the detection) and one for the per-voxel bounding cylinder regression.

With the classification branch we want to find the voxel which should be used for the bounding cylinder regression.
This is done by classifying each voxel into one of two classes: containing the top of a cylinder or not.
At the moment we assume that each voxel can only have one bounding cylinder.
This could be extended by having several bounding cylinder per voxel, or a second network for refinement.

Since there are not many point cloud datasets where the instance segmentation and/or detection are provided together with the 3D joints of each instance we decided to work on datasets that provide at least the point cloud and the joint positions. The cylinders are then defined based on the joint positions for each person.

We define each cylinder axis as being aligned with the person's neck. The top of the cylinder is at the same height as the joint which is furthest from the ground (for this person). The radius of the cylinder is determined by the distance of the joint furthest from the neck axis of this person.

During training we use the ground truth of the classification to mask the voxel that should be used for back propagation while during testing and inference time we use the output of the classification branch to retrieve the voxel which contains the desired bounding cylinders, as in the RCNN family of frameworks.

The loss for this part is defined as: 
\begin{equation}
\begin{split}
L(\vec{v}, \vec{b}) = \;& \frac{1}{N_{cls}} \sum\limits_{i} L_{cls}(v_{i}, v_{i}^*) \; + \\
                        & \lambda \frac{1}{N_{reg}} \sum\limits_{i}  v_i^* L_{reg}(b_{i}, b_{i}^*)
\end{split}
\end{equation}

Here $\vec{v}$ represents the output of the head of our network doing the classification and $\vec{b}$ represents the output of the head doing the bounding cylinder regression. Their respective ground truth is represented by $\vec{v^*}$ and $\vec{b^*}$.
$L_{cls}$ is a cross-entropy loss doing the classification of voxels.
For the regression loss, we use $L_{reg}(b_i,b_{i^*}) = R(b_i - b_{i^*})$ where $R$ is the robust loss function (smooth L1) defined in \cite{girshick2015fastrcnn}.

In the case of human detection only a relatively small number of voxels contain a cylinder to regress, so we need to re-sample our data.
Instead of working with $\vec{v}$ and $\vec{b}$ which is the entire output of our networks, we use every voxel containing an object across the batch, as many non empty voxels randomly chosen across the batch and the same amount of voxel across every voxel in the batch.
In the case where there is no human in our whole batch, we just pick 32 random voxels. 
That way, we maintain a bias towards "empty" voxels, we are sure to use the voxels with data in them and we do not have to compute the whole loss on every empty voxel in our batch.

The two terms of the loss are normalized by $N_{cls}$ and $N_{reg}$ and weighted by a balancing parameter $\lambda$. 
In our current implementation, $N_{cls}$ and $N_{reg}$ both set to the number of voxels used after the sampling while $\lambda$ is set to 1.

We use a normalization analog to the one presented in \cite[Equation 1]{zhou2017voxelnet}, \ie the cylinder coordinates are normalized relative to the voxel size. By transforming the cylinders back into our world coordinate system we can extract a point cloud consisting of all points within this cylinder.

% Suppose we look at cylinder $i$ in our scene and lets define $d_{xz}$ and $d_{xy}$ as being the diagonal size of a voxel on the $xz$ and $xy$ plane.
% We denote $\tilde{\vec{b}}^*$ as being the ground truth cylinders in our euclidean space and $\tilde{\vec{b}}$ as the re-projection of the output of our networks into our 3D euclidean space and $r$ being the radius of a bounding cylinder and $\vec{a}$ being the voxel in which the top of the cylinder is.
% Then, we use the following parametrization defined in \cite{zhou2017voxelnet} and change it a bit for our use case:

% \begin{align}
% b^{i}_{x} = \frac{\tilde{b}^{i}_{x} - a^{i}_{x}}{d_{xz}} \quad b^{*i}_{x} = \frac{\tilde{b}^{*i}_{x} - a^{i}_{x}}{d_{xz}} \\
% b^{i}_{y} = \frac{\tilde{b}^{i}_{y} - a^{i}_{y}}{d_{xy}}\quad b^{*i}_{y} = \frac{\tilde{b}^{*i}_{y} - a^{i}_{y}}{d_{xy}}\\
% b^{i}_{z} = \frac{\tilde{b}^{i}_{z} - a^{i}_{z}}{d_{xz}} \quad b^{*i}_{z} = \frac{\tilde{b}^{*i}_{z} - a^{i}_{z}}{d_{xz}}\\
% b^{i}_{r} = \log{\tilde{b}^{i}_{r}} \quad\quad b^{*i}_{r} = \log{\tilde{b}^{*i}_{r}} \quad
% \end{align}

% By reverting the parametrization, we can then get bounding cylinder in the original Euclidean space, which we can  then be used to extract for each detected human the corresponding sub point cloud.
% At the moment, we consider that every person in the scene are in contact with the ground.

\subsection{Instance wise processing}
\begin{figure}[htb]
    \includegraphics[width=\linewidth]{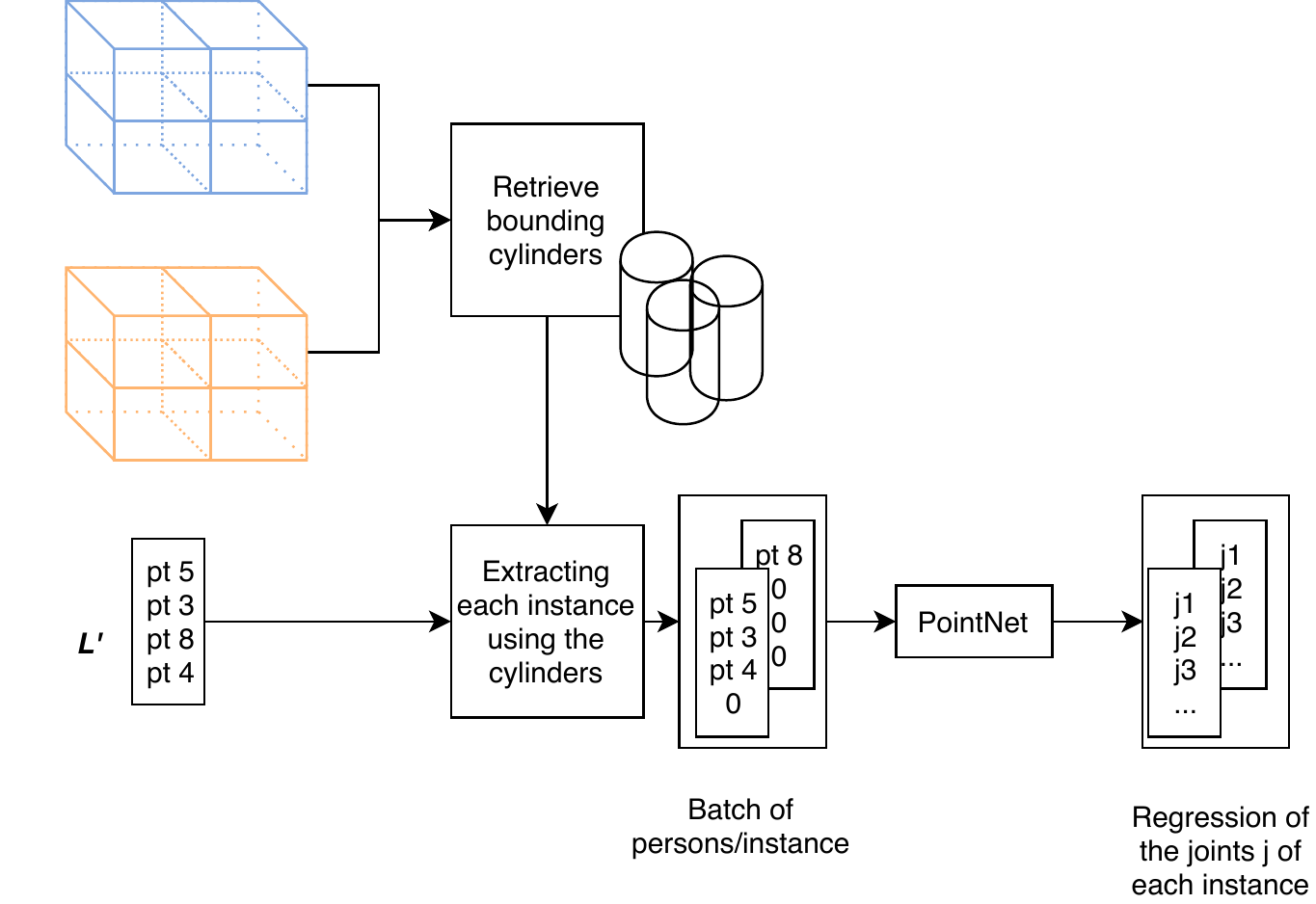}
    \caption{In this last part, we use the predictions from our two branches to retrieve the bounding cylinder and extract the point cloud of each instance. These points clouds are then fed into a PointNet \cite{qi2017pointnet} in order to regress the joints.}
  \label{fig:joints_reg}
\end{figure}
Using the extracted point cloud for each cylinder, we can set up a batch of instances to be processed by the last stage of our framework. During training we take at most 32 persons across the frames in the minibatch and for each person we sample at most 1024 points. If a person does not have enough points (32 in our case), we discard this instance.

If there are no extracted point clouds at all, \ie there was no detection at all, we skip this whole part and just compute the loss previously mentioned.
In our use case, we are doing joint regression from the point cloud using PointNet \cite{qi2017pointnet}.%%a modified
 %where we replaced the simple fully connected layers by some dense blocks composed of convolution with a kernel size 1.
%By using those instead of a simple convolution, the dense block should learn different projections (as many projections as the growing rate) in a given space and will try then to combined those projections to project to a space of a higher dimension. The detail of this architecture can be found in the supplementary material. We will investigate this architecture in a later work, since this is not the main goal of this paper.
This framework could of course be modified to replace what should be regressed and how it is regressed.
The only normalization we do on those point cloud is to put them in a sphere located at the middle of cylinder with a radius of half the height of the cylinder.
The final loss of our whole framework is:
\begin{equation}
\begin{split}
L(\vec{v}, \vec{b}, \vec{j})= \;& \frac{1}{N_{cls}} \sum\limits_{i} L_{cls}(v_{i}, v_{i}^*) \; + \\
            & \lambda \frac{1}{N_{reg}} \sum\limits_{i}  v_i^* L_{reg}(b_{i}, b_{i}^*) + \\
            & \lambda_2 \frac{1}{N_{joints}} \sum\limits_{k=0}^{N_{inst}} L_{joints\_reg}(j_{k}, j_{k}^*)
\end{split}
\end{equation}

Here, we take the loss of the first part and add the loss of the regression part.
The loss for the joint regression $L_{joints\_reg}$ we use is Mean Square Error Loss calculated between the joints predicted $\vec{j}$ and the ground truth joints $\vec{j}*$. As stated before, we sample $N_{inst}$ through our batch for the regression part, this $N_{inst}$ being 32 during training.

If there are too many persons we pick the person to regress randomly across the batch. If there are not enough persons, we pad this batch with zeros. During testing, we use every detected bounding box instead. We add another balancing term $\lambda_2$ that we set to 1, $\lambda$ being set to 1 too.

The $N_{joints}$ is the total number of joints to regress across our $N_{inst}$ instances.
We show that using an end to end training on this whole framework improves the result compared to a two-stage solution trained separately.
%We investigate how using the feature of the whole scene could benefit the joints %regression. 
%In order to use the feature from the whole scene, we take the per-voxel feature of our %network for each point, we concatenate the feature from the corresponding voxel.

\section{Experiments}
To simulate real world scenarios we designed four challenging experiments and compared our results with the baseline method. In particular, we train stage-wise state-of-the-art 3D point cloud based object detection via VoxelNet\cite{zhou2017voxelnet} and a state-of-the-art point cloud based regression network, namely PointNet\cite{qi2017pointnet}.

To the best of our knowledge no end-to-end solution is available for pure point cloud based 3D multi-person pose estimation. Thus for comparison we cascade the above mentioned state-of-the-art algorithms to perform each part, \ie 3D person detection and per-person joint detection. Each part is trained separately and the best model is chosen for final evaluation. 

In order to simulate camera failures in real life scenarios all datasets used in first three experiments are conducted with a random number of camera inputs, \ie we randomly drop one or more of the camera inputs. Details are discussed in the respective experiment sections.

The first three experiments are conducted on the CMU Panoptic dataset\cite{joo2017pano}, which was created to capture social interaction using several modalities.
Here several actors are interacting in a closed environment and captured by a multitude of RGB and RGBD cameras. For our experiments we use only the depth images from the ten Kinect 2 RGBD sensors and show that point clouds alone are sufficient to accurately detect people and identify their pose.

The cameras in the CMU Panoptic dataset are placed in various positions on a dome over the room, all pointing towards the center. The dataset contains several different social scenes with 3D joint annotations of the actors. For our experiments we randomly choose four scenes to conduct our evaluation. Namely, ``160224\_haggling1'', ``160226\_haggling1'', ``160422\_haggling1'', ``160906\_pizza1''. For more information about the dataset, readers are referred to \cite{joo2017pano}. 

The last experiment is conducted on the MVOR dataset \cite{srivastav2018mvor}, which was acquired over 4 days at a hospital operation room with a varying number of medical personnel visible in the scene and recorded by three RGBD cameras. The dataset captures a real world operation room scenarios, including cluttered background, various medical devices, random number of people at any given time, and ubiquitous occlusion caused by medical devices. The point clouds were cropped to a 4x2x3 m size cube and sampled using a voxel grid filter with a 2.5x2.5x2.5 cm grid size.

\subsection{Metrics}
To evaluate the overall performance of the algorithm we use both 3D object detection metrics, Average Precision (AP) on Intersection of Union (IoU) > 0.5, used in KITTI Vision Benchmark Suite \cite{Geiger2012CVPR} to evaluate the instance detection part of the algorithm, and per joint mean distance metrics, denoted as DIST for per joint distance and ACC for accuracy under 10cm threshold in the following sections. When calculating the distance, we only count the joints in true positive detections. And if there is duplicated detections, we only calculate the joints difference of the highest scored detection. 

\subsection{View Generalization}
In this experiment, we simulate cameras being placed in different locations in the room to demonstrate the view generalization of the algorithm. We achieve this by using different cameras at training and test time. We show that our algorithm is robust when changing the view or location of the camera.

For our training dataset we randomly choose 8 camera inputs as our training dataset, namely camera 0, 1, 2, 3, 4, 6, 7, 9. The remaining cameras 5 and 8 are used for testing. Note that in both training and testing, we choose a random number of cameras at any given time to simulate camera failures. 

The first experiment is conducted on scene ``160224\_haggling1'', ``160226\_haggling1'' and ``160906\_pizza1''. The training dataset has 6858 frames, uniformly down-sampled by a factor of 3. The testing dataset has 897 frames, and is down-sampled by a factor of 30.

As we can see from \autoref{tab_exp1}, even with severe camera view changes and camera failures, the proposed method performs well. Meanwhile, our end-to-end solution outperforms the baseline by a large margin, both from detection perspective and joint regression perspective.  

\begin{table}[ht]
\centering
\begin{tabular}{lccccc}
\toprule
           & \multicolumn{2}{c}{{[}1{]} + {[}16{]}} & & \multicolumn{2}{c}{Point R-CNN} \\
\cmidrule{2-3} \cmidrule{5-6}
           & DIST & ACC & & DIST & ACC \\
           & [cm] & \%  & & [cm] & \% \\
\midrule
Neck       & 5.47  & 94.8 && \textbf{4.67}  & \textbf{96.7} \\
Headtop    & 10.55 & 52.1 && \textbf{9.28}  & \textbf{63.5} \\
BodyCenter & 7.14  & 85.5 && \textbf{6.89}  & \textbf{85.6} \\
Lshoulder  & 19.00 & 5.2  && \textbf{17.00} &\textbf{ 22.9} \\
% Lelbow     & 24.41 & 1.8  && 22.03 & 15.7 \\
% Lwrist     & 27.55 & 1.0  && 25.98 & 3.1  \\
Lhip       & 11.55 & 40.4 && \textbf{10.6}1 & \textbf{52.8} \\
Lknee      & 13.41 & 33.8 && \textbf{12.80} & \textbf{40.8} \\
Lankle     & 16.36 & 28.3 && \textbf{15.75} & \textbf{30.5} \\
Rshoulder    & 18.16 & 3.4  && \textbf{16.28} & \textbf{25.7} \\
% Relbow     & 25.18 & 0.6  && 23.13 & 12.8 \\
% Rwrist     & 28.03 & 0.9  && 26.93 & 6.0  \\
Rhip       & 12.29 & 32.1 && \textbf{11.42} & \textbf{45.6} \\
Rknee      & 14.41 & 24.7 && \textbf{13.52} & \textbf{34.8} \\
Rankle     & 17.05 & 24.4 &&\textbf{ 15.95 }& \textbf{28.2} \\
\midrule
Mean       & 13.20 & 38.6 && \textbf{12.20} & \textbf{47.9} \\
\midrule
AP         &        \multicolumn{2}{c}{81.24}& & \multicolumn{2}{c}{\textbf{81.37}}\\

\bottomrule 
\end{tabular}
\caption{View generalization: per joint distance in cm and accuracy < 10 cm.}
\label{tab_exp1}
\end{table}

%To test our framework and our claims, we conduct 4 experiments. The 3 first experiments are done on the CMU Panoptic dataset and have for goal to (1) demonstrate the view generalization, (2) actor generalization and both generalization at the same time (3). The fourth experiment is done on the

\subsection{Actor generalization}
In this experiment, we explore the generalization problem in terms of the objects and scenes. In particular, we train the network with scene ``160224\_haggling1'', ``160226\_haggling1'' and ``160906\_pizza1'' and tested with scene ``160422\_haggling1'' with same camera view, \ie both training and testing data uses camera  0, 1, 2, 3, 4, 6, 7, 9 with random camera failure. The training dataset has 6858 frames, uniformly downsampled by factor 3. The testing dataset has 430 frames, down-sampled with factor 30. \autoref{tab_exp2} shows that the proposed method outperforms the baseline method by a large margin, especially for shoulders, hips and Knees.

\begin{table}[]
\centering
\begin{tabular}{lccccc}
\toprule
           & \multicolumn{2}{c}{{[}1{]} + {[}16{]}} & & \multicolumn{2}{c}{Point R-CNN} \\
\cmidrule{2-3} \cmidrule{5-6}
           & DIST & ACC & & DIST & ACC \\
           & [cm] & \%  & & [cm] & \% \\
\midrule
Neck       & 7.28  & 81.0 && \textbf{5.65}  & \textbf{89.3} \\
Headtop    & 12.10 & 38.2 && \textbf{8.40 } & \textbf{75.1} \\
BodyCenter & 7.24  & 81.7 && \textbf{6.50}  & \textbf{86.5} \\
Lshoulder  & 19.83 & 0.4  && \textbf{12.96} & \textbf{42.7} \\
% Lelbow     & 24.80 & 0.8  && 16.97 & 30.5 \\
% Lwrist     & 25.91 & 3.8  && 23.20 & 7.7  \\
Lhip       & 12.26 & 28.8 && \textbf{9.24}  & \textbf{63.6 }\\
Lknee      & 13.12 & 30.3 && \textbf{10.23} & \textbf{56.6} \\
Lankle     & 16.49 & 25.8 && \textbf{13.67} & \textbf{37.4} \\
Rshoulder    & 19.00 & 5.4  && \textbf{12.49} & \textbf{44.8} \\
% Relbow     & 24.49 & 2.1  && 16.07 & 28.8 \\
% Rwrist     & 25.17 & 1.8  && 21.33 & 16.6 \\
Rhip       & 12.85 & 27.8 && \textbf{9.42}  & \textbf{64.4} \\
Rknee      & 13.95 & 27.2 && \textbf{10.22} & \textbf{56.6} \\
Rankle     & 17.71 & 17.6 && \textbf{13.82} & \textbf{36.3} \\
\midrule
Mean       & 13.80 & 33.1 && \textbf{10.24} & \textbf{59.4} \\
\midrule
AP         &        \multicolumn{2}{c}{80.65}& & \multicolumn{2}{c}{\textbf{90.54}}\\
\bottomrule
\end{tabular}
\caption{Actor generalization: per joint distance in cm and accuracy < 10cm. }
\label{tab_exp2}
\end{table}

\subsection{View and Actor generalization}
In this study, we demonstrate the evaluation under both view change and actor changes. We trained the network with scene  ``160224\_haggling1``, ``160226\_haggling1'' and ``160906\_pizza1'' using camera 0, 1, 2, 3, 4, 6, 7, 9 and tested on scene ``160422\_haggling1'' on camera 5, 8, both with random camera failures. The training dataset has 6858 frames, uniformly downsampled with factor 3. Testing dataset has 432 frames, downsampled with factor 30. From \autoref{tab_exp3} we can see that the proposed method outperforms in all the joints, some with large margins, for example the accuracy of headtop, left and right shoulder joints.

\begin{table}
\centering
\begin{tabular}{lccccc}
\toprule
           & \multicolumn{2}{c}{{[}1{]} + {[}16{]}} & & \multicolumn{2}{c}{Point R-CNN} \\
\cmidrule{2-3} \cmidrule{5-6}
           & DIST & ACC & & DIST & ACC \\
           & [cm] & \%  & & [cm] & \% \\
\midrule
Neck       & 7.88  & 71.1 && \textbf{6.32}  & \textbf{88.4} \\
Headtop    & 9.99  & 54.9 && \textbf{8.64}  & \textbf{71.5} \\
BodyCenter & 7.20  & 80.2 && \textbf{6.29}  & \textbf{87.0} \\
Lshoulder  & 13.57 & 37.2 && \textbf{11.50} & \textbf{48.7} \\
% Lelbow     & 17.56 & 23.3 && 15.41 & 33.0 \\
% Lwrist     & 23.97 & 7.5  && 22.12 & 7.8  \\
Lhip       & 9.18  & 64.8 && \textbf{8.22}  & \textbf{73.6} \\
Lknee      & 9.28  & 63.2 && \textbf{8.12} & \textbf{74.1} \\
Lankle     & 12.15 & 45.9 && \textbf{10.99} & \textbf{52.7} \\
Rshoulder    & 14.37 & 31.5 && \textbf{11.71} & \textbf{48.1} \\
% Relbow     & 17.87 & 21.6 && 15.33 & 31.1 \\
% Rwrist     & 22.87 & 14.1 && 20.89 & 15.9 \\
Rhip       & 10.51 & 53.3 && \textbf{8.77}  & \textbf{69.5} \\
Rknee      & 11.29 & 51.0 && \textbf{9.49}  & \textbf{64.1} \\
Rankle     & 14.43 & 35.9 && \textbf{12.62} & \textbf{44.2} \\
\midrule
Mean       & 10.90 & 53.6 &&\textbf{9.33} & \textbf{65.64}\\
\midrule
AP         &        \multicolumn{2}{c}{71.25}& & \multicolumn{2}{c}{\textbf{89.94}}\\
\bottomrule
\end{tabular}
\caption{View and actor generalization: per joint distance in cm and accuracy < 10cm. }
\label{tab_exp3}
\end{table}

\subsection{Handling background and cluttered scenes}
In this experiment, we use the MVOR dataset which shows a very cluttered room and walls in the background. Unlike previous datasets, this is a real world operation room scenario, which is naturally more complex. Furthermore, due to the size of the dataset, it is not feasible to train the network from scratch, as it is mentioned in the original paper \cite{mvor}, so in this experiment, we fine-tuned our previous model on the first 3 days of data and tested on the 4th day data with all 3 cameras used. Since the number of joints is different from the previous dataset, we only fine-tuned the joints common to both dataset's annotations. The training dataset has 513 frames and 113 testing frames. \autoref{tab_exp4} shows that our algorithm outperforms the baseline.

\begin{table}
\centering
\begin{tabular}{lccccc}
\toprule
           & \multicolumn{2}{c}{{[}1{]} + {[}16{]}} & & \multicolumn{2}{c}{Point R-CNN} \\
\cmidrule{2-3} \cmidrule{5-6}
           & DIST & ACC & & DIST & ACC \\
           & [cm] & \%  & & [cm] & \% \\
\midrule
Head    & 13.06 & 37.5 && \textbf{9.74}  & \textbf{61.9} \\
Neck    & 12.05 & 41.3 && \textbf{9.24}  & \textbf{65.0} \\
Lshoulder & 19.02 & 15.3 && \textbf{15.52} & \textbf{28.9} \\
Rshoulder & 16.90 & 21.0 && \textbf{15.12} & \textbf{27.4} \\
Lhip    & 27.93 & 7.5  && \textbf{20.27} & \textbf{14.2} \\
Rhip    & 32.34 & 8.2  && \textbf{18.50} & \textbf{19.3} \\
Lelb    & 27.60 & 6.2  && \textbf{22.03} & \textbf{12.7} \\
Relb    & 25.87 & 6.1  && \textbf{20.41} & \textbf{14.2} \\
% Lwrist  & 29.32 & 5.7  && 22.77 & 10.2 \\
% Rwist   & 30.94 & 5.2  && 23.38 & 7.1  \\
\midrule
Mean    & 24.96 & 15.4 && \textbf{17.70} & \textbf{26.1}\\
\midrule
AP         &        \multicolumn{2}{c}{57.84}& & \multicolumn{2}{c}{\textbf{64.73}}\\
\bottomrule
\end{tabular}
\caption{MVOR dataset: per joint distance in cm and accuracy < 10cm. }
\label{tab_exp4}
\end{table}

\begin{figure}[ht]
\centering
\includegraphics[width=\linewidth]{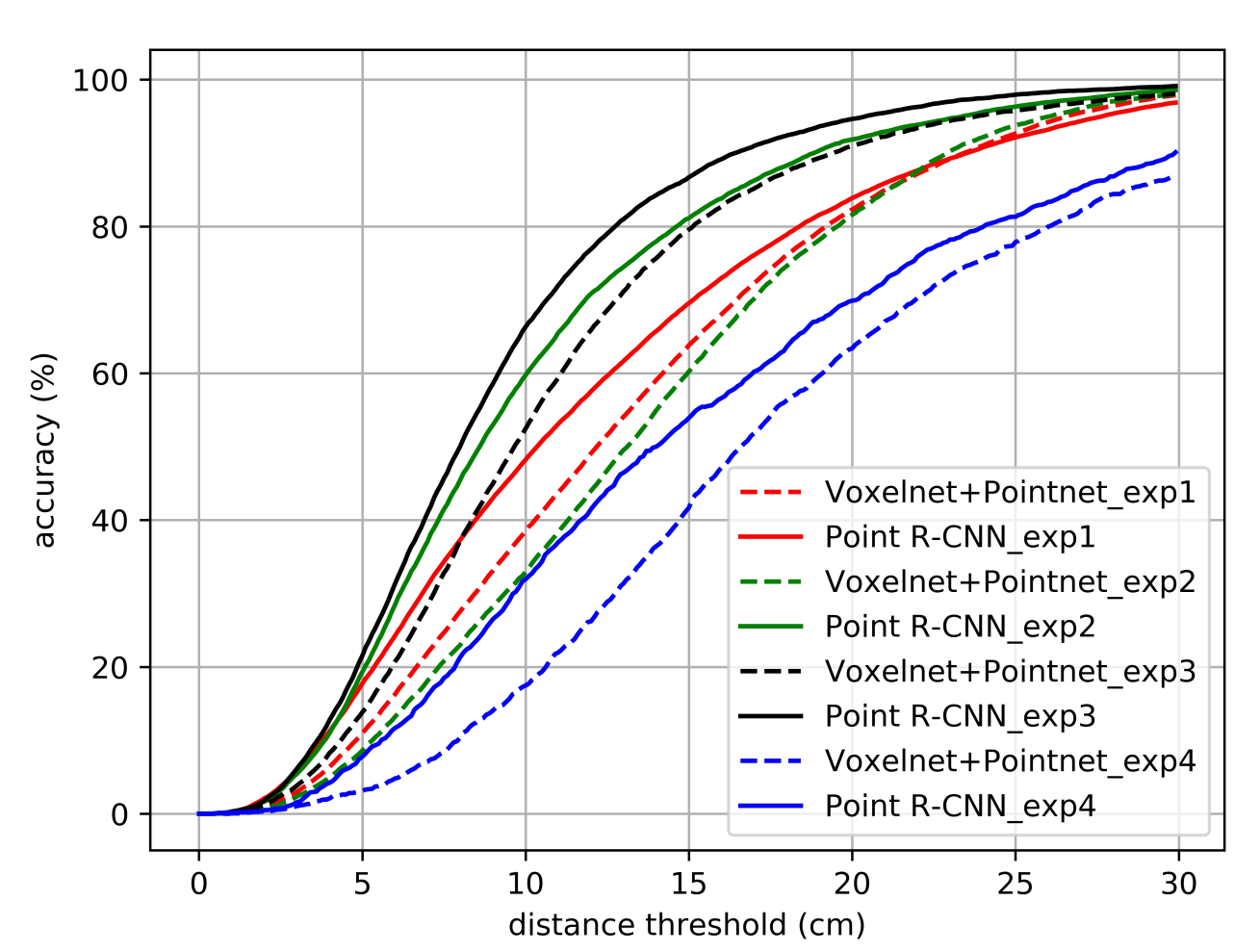}
\caption{The mean per joint distance accuracy using different thresholds for the above four experiments. Each dotted line represents the baseline method and the solid line represents the proposed method. Experiments are color coded. }
\label{fig:img_plot}
\end{figure}

\begin{figure*}[ht]
\centering
    \includegraphics[width=\linewidth]{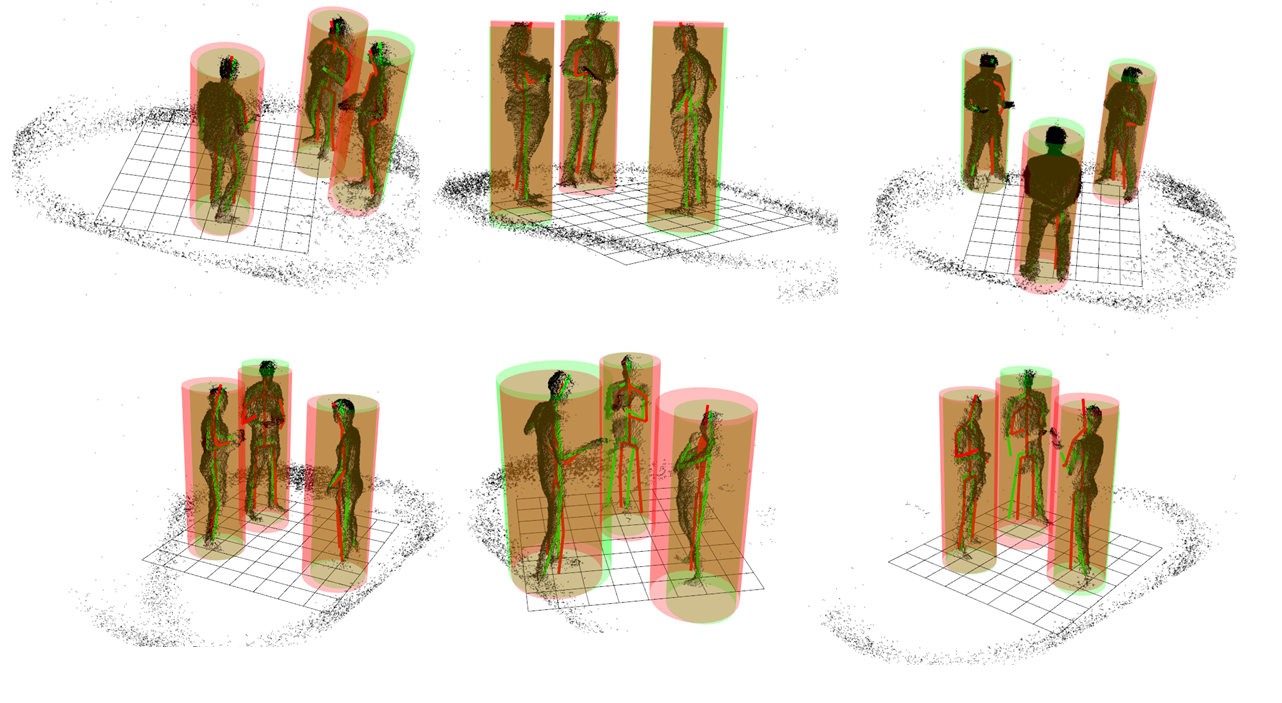}
    \includegraphics[width=\linewidth]{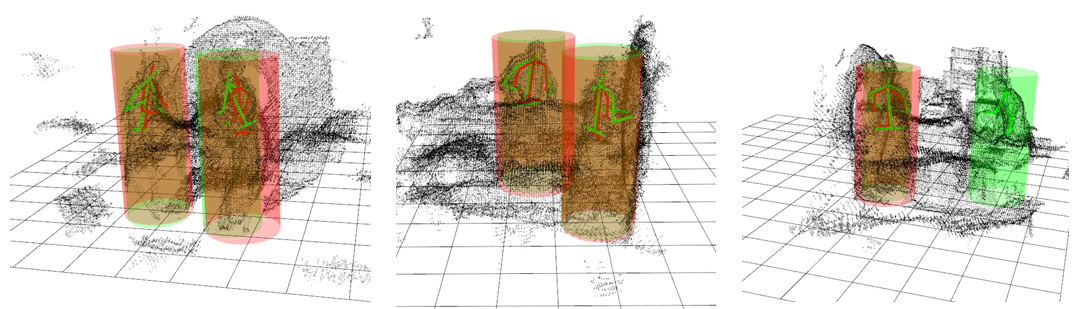}
  \caption{Person detection and pose estimation results on CMU Panoptic (rows 1, 2) and MVOR (row 3). The red cylinder and skeleton are the prediction result and the green cylinder and skeleton indicate the ground truth.}
  \label{fig:img_exp1}
\end{figure*}

\subsection{Qualitative Evaluation and Discussion}
Apart from the experiments we mentioned above, we evaluated the mean of per joint distance accuracy under different thresholds. As shown in \autoref{fig:img_plot}, the proposed method consistently outperforms the baseline method under various thresholds in all four experiments. Sample testing results from all experiments are shown in \autoref{fig:img_exp1}, where the ground truth is color coded green and the prediction red.

\section{Conclusion}

In this work we have demonstrated through extended experiments that point cloud is the natural and straightforward alternative for a multi-sensor indoor system, where the fusion of multi-sensor information is efficient. Unlike conventional methods that use complex fusion models to combine information, which tend to generalize poorly, we show through various challenging real world scenarios that the proposed algorithm can generalize well. Furthermore, we propose an end-to-end multi-person 3D pose estimation network, ``Point R-CNN'', and show that the proposed network outperforms the simply cascaded model by large margins in various experiments. The study shows that using an end-to-end network greatly improves both object detection and joint regression performance.

%\afterpage{\clearpage}
\FloatBarrier

\bibliographystyle{unsrt}
\bibliography{cvpr_2019}

\begin{thebibliography}{10}

\bibitem{zhou2017voxelnet}
Yin Zhou and Oncel Tuzel.
\newblock Voxel{N}et: End-to-end learning for point cloud based 3d object
  detection.
\newblock {\em arXiv preprint arXiv:1711.06396}, 2017.

\bibitem{qi2017frustum}
Charles~R Qi, Wei Liu, Chenxia Wu, Hao Su, and Leonidas~J Guibas.
\newblock Frustum pointnets for 3d object detection from rgb-d data.
\newblock {\em arXiv preprint arXiv:1711.08488}, 2017.

\bibitem{ku2017joint}
Jason Ku, Melissa Mozifian, Jungwook Lee, Ali Harakeh, and Steven Waslander.
\newblock Joint 3d proposal generation and object detection from view
  aggregation.
\newblock {\em arXiv preprint arXiv:1712.02294}, 2017.

\bibitem{park2008multiple}
Youngmin Park, Vincent Lepetit, and Woontack Woo.
\newblock Multiple 3d object tracking for augmented reality.
\newblock In {\em Proceedings of the 7th IEEE/ACM International Symposium on
  Mixed and Augmented Reality}, pages 117--120. IEEE Computer Society, 2008.

\bibitem{singh2017darwin}
Vivek Singh, Kai Ma, Birgi Tamersoy, Yao-Jen Chang, Andreas Wimmer, Thomas
  O’Donnell, and Terrence Chen.
\newblock Darwin: Deformable patient avatar representation with deep image
  network.
\newblock In {\em International Conference on Medical Image Computing and
  Computer-Assisted Intervention}, pages 497--504. Springer, 2017.

\bibitem{kadkhodamohammadi2017multi}
Abdolrahim Kadkhodamohammadi, Afshin Gangi, Michel de~Mathelin, and Nicolas
  Padoy.
\newblock A multi-view rgb-d approach for human pose estimation in operating
  rooms.
\newblock In {\em Applications of Computer Vision (WACV), 2017 IEEE Winter
  Conference on}, pages 363--372. IEEE, 2017.

\bibitem{joo2018total}
Hanbyul Joo, Tomas Simon, and Yaser Sheikh.
\newblock Total capture: A 3d deformation model for tracking faces, hands, and
  bodies.
\newblock In {\em Proceedings of the IEEE Conference on Computer Vision and
  Pattern Recognition}, pages 8320--8329, 2018.

\bibitem{cho2014multi}
Hyunggi Cho, Young-Woo Seo, BVK~Vijaya Kumar, and Ragunathan~Raj Rajkumar.
\newblock A multi-sensor fusion system for moving object detection and tracking
  in urban driving environments.
\newblock In {\em Robotics and Automation (ICRA), 2014 IEEE International
  Conference on}, pages 1836--1843. IEEE, 2014.

\bibitem{gupta2015aligning}
Saurabh Gupta, Pablo Arbel{\'a}ez, Ross Girshick, and Jitendra Malik.
\newblock Aligning 3d models to rgb-d images of cluttered scenes.
\newblock In {\em Proceedings of the IEEE Conference on Computer Vision and
  Pattern Recognition}, pages 4731--4740, 2015.

\bibitem{tulsiani2015viewpoints}
Shubham Tulsiani and Jitendra Malik.
\newblock Viewpoints and keypoints.
\newblock In {\em Proceedings of the IEEE Conference on Computer Vision and
  Pattern Recognition}, pages 1510--1519, 2015.

\bibitem{xu2017pointfusion}
Danfei Xu, Dragomir Anguelov, and Ashesh Jain.
\newblock Pointfusion: Deep sensor fusion for 3d bounding box estimation.
\newblock {\em arXiv preprint arXiv:1711.10871}, 2017.

\bibitem{mvor}
Vinkle Srivastav, Thibaut Issenhuth, Abdolrahim Kadkhodamohammadi, Michel~de
  Mathelin, Afshin Gangi, and Nicolas Padoy.
\newblock Mvor: A multi-view rgb-d operating room dataset for 2d and 3d human
  pose estimation.
\newblock {\em arXiv preprint arXiv:1808.08180}, 2018.

\bibitem{wu20153d}
Zhirong Wu, Shuran Song, Aditya Khosla, Fisher Yu, Linguang Zhang, Xiaoou Tang,
  and Jianxiong Xiao.
\newblock 3d shapenets: A deep representation for volumetric shapes.
\newblock In {\em Proceedings of the IEEE conference on computer vision and
  pattern recognition}, pages 1912--1920, 2015.

\bibitem{riegler2017octnet}
Gernot Riegler, Ali~Osman Ulusoy, and Andreas Geiger.
\newblock Octnet: Learning deep 3d representations at high resolutions.
\newblock In {\em Proceedings of the IEEE Conference on Computer Vision and
  Pattern Recognition}, volume~3, 2017.

\bibitem{jampani2016learning}
Varun Jampani, Martin Kiefel, and Peter~V Gehler.
\newblock Learning sparse high dimensional filters: Image filtering, dense crfs
  and bilateral neural networks.
\newblock In {\em Proceedings of the IEEE Conference on Computer Vision and
  Pattern Recognition}, pages 4452--4461, 2016.

\bibitem{qi2017pointnet}
Charles~R Qi, Hao Su, Kaichun Mo, and Leonidas~J Guibas.
\newblock Pointnet: Deep learning on point sets for 3d classification and
  segmentation.
\newblock {\em Proc. Computer Vision and Pattern Recognition (CVPR), IEEE},
  1(2):4, 2017.

\bibitem{qi2017pointnet++}
Charles~Ruizhongtai Qi, Li~Yi, Hao Su, and Leonidas~J Guibas.
\newblock Pointnet++: Deep hierarchical feature learning on point sets in a
  metric space.
\newblock In {\em Advances in Neural Information Processing Systems}, pages
  5099--5108, 2017.

\bibitem{ge2018hand}
Liuhao Ge, Yujun Cai, Junwu Weng, and Junsong Yuan.
\newblock Hand pointnet: 3d hand pose estimation using point sets.
\newblock In {\em Proceedings of the IEEE Conference on Computer Vision and
  Pattern Recognition}, pages 8417--8426, 2018.

\bibitem{su2018splatnet}
Hang Su, Varun Jampani, Deqing Sun, Subhransu Maji, Evangelos Kalogerakis,
  Ming-Hsuan Yang, and Jan Kautz.
\newblock Splatnet: Sparse lattice networks for point cloud processing.
\newblock In {\em Proceedings of the IEEE Conference on Computer Vision and
  Pattern Recognition}, pages 2530--2539, 2018.

\bibitem{li2018pointcnn}
Yangyan Li, Rui Bu, Mingchao Sun, and Baoquan Chen.
\newblock Pointcnn.
\newblock {\em arXiv preprint arXiv:1801.07791}, 2018.

\bibitem{hegde2016fusionnet}
Vishakh Hegde and Reza Zadeh.
\newblock Fusionnet: 3d object classification using multiple data
  representations.
\newblock {\em arXiv preprint arXiv:1607.05695}, 2016.

\bibitem{sarafianos20163d}
Nikolaos Sarafianos, Bogdan Boteanu, Bogdan Ionescu, and Ioannis~A Kakadiaris.
\newblock 3d human pose estimation: A review of the literature and analysis of
  covariates.
\newblock {\em Computer Vision and Image Understanding}, 152:1--20, 2016.

\bibitem{yang20183d}
Wei Yang, Wanli Ouyang, Xiaolong Wang, Jimmy Ren, Hongsheng Li, and Xiaogang
  Wang.
\newblock 3d human pose estimation in the wild by adversarial learning.
\newblock In {\em Proceedings of the IEEE Conference on Computer Vision and
  Pattern Recognition}, volume~1, 2018.

\bibitem{rhodin2018learning}
Helge Rhodin, J{\"o}rg Sp{\"o}rri, Isinsu Katircioglu, Victor Constantin,
  Fr{\'e}d{\'e}ric Meyer, Erich M{\"u}ller, Mathieu Salzmann, and Pascal Fua.
\newblock Learning monocular 3d human pose estimation from multi-view images.
\newblock In {\em Proceedings of the IEEE Conference on Computer Vision and
  Pattern Recognition}, 2018.

\bibitem{kadkhodamohammadi2018generalizable}
Abdolrahim Kadkhodamohammadi and Nicolas Padoy.
\newblock A generalizable approach for multi-view 3d human pose regression.
\newblock {\em arXiv preprint arXiv:1804.10462}, 2018.

\bibitem{moon2018v2v}
Gyeongsik Moon, Ju~Yong Chang, and Kyoung~Mu Lee.
\newblock V2v-posenet: Voxel-to-voxel prediction network for accurate 3d hand
  and human pose estimation from a single depth map.
\newblock In {\em Proceedings of the IEEE Conference on Computer Vision and
  Pattern Recognition}, volume~2, 2018.

\bibitem{haque2016pose3d}
Albert Haque, Boya Peng, Zelun Luo, Alexandre Alahi, Serena Yeung, and
  Fei{-}Fei Li.
\newblock Viewpoint invariant 3d human pose estimation with recurrent error
  feedback.
\newblock {\em CoRR}, abs/1603.07076, 2016.

\bibitem{girshick2015fastrcnn}
Ross~B. Girshick.
\newblock Fast {R-CNN}.
\newblock {\em CoRR}, abs/1504.08083, 2015.

\bibitem{he2017maskrcnn}
Kaiming He, Georgia Gkioxari, Piotr Doll{\'{a}}r, and Ross~B. Girshick.
\newblock Mask {R-CNN}.
\newblock {\em CoRR}, abs/1703.06870, 2017.

\bibitem{denseunet}
Xiaomeng Li, Hao Chen, Xiaojuan Qi, Qi~Dou, Chi{-}Wing Fu, and Pheng{-}Ann
  Heng.
\newblock H-denseunet: Hybrid densely connected unet for liver and liver tumor
  segmentation from {CT} volumes.
\newblock {\em CoRR}, abs/1709.07330, 2017.

\bibitem{joo2017pano}
Hanbyul Joo, Tomas Simon, Xulong Li, Hao Liu, Lei Tan, Lin Gui, Sean Banerjee,
  Timothy~Scott Godisart, Bart Nabbe, Iain Matthews, Takeo Kanade, Shohei
  Nobuhara, and Yaser Sheikh.
\newblock Panoptic studio: A massively multiview system for social interaction
  capture.
\newblock {\em IEEE Transactions on Pattern Analysis and Machine Intelligence},
  2017.

\bibitem{srivastav2018mvor}
V.~{Srivastav}, T.~{Issenhuth}, A.~{Kadkhodamohammadi}, M.~{de Mathelin},
  A.~{Gangi}, and N.~{Padoy}.
\newblock {MVOR: A Multi-view RGB-D Operating Room Dataset for 2D and 3D Human
  Pose Estimation}.
\newblock {\em ArXiv e-prints}, August 2018.

\bibitem{Geiger2012CVPR}
Andreas Geiger, Philip Lenz, and Raquel Urtasun.
\newblock Are we ready for autonomous driving? the kitti vision benchmark
  suite.
\newblock In {\em Conference on Computer Vision and Pattern Recognition
  (CVPR)}, 2012.

\end{thebibliography}

\end{document}